\documentclass[conference]{IEEEtran}
\IEEEoverridecommandlockouts
\usepackage{graphicx}
\usepackage{amsmath,amssymb,amsthm} 
\usepackage{color}
\usepackage{multirow}
\usepackage[flushleft]{threeparttable} 
\usepackage{boldline} 
\usepackage[table]{xcolor} 
\usepackage{algorithm,algorithmic,bm,color}
\usepackage[colorlinks=true,allcolors=black]{hyperref}
\usepackage[numbers,sort&compress]{natbib}

\newcommand{\argmin}{\arg\!\min}

\newcommand{\Gmat}[0]{{{\bf G}}}
\newcommand{\Hmat}[0]{{{\bf H}}}
\newcommand{\Imat}{{\bf I}}

\newcommand{\Rmat}[0]{{{\bf R}}}
\newcommand{\Smat}[0]{{{\bf S}}}

\newcommand{\Xmat}{{\bf X}}

\def\BibTeX{{\rm B\kern-.05em{\sc i\kern-.025em b}\kern-.08em
    T\kern-.1667em\lower.7ex\hbox{E}\kern-.125emX}}
\begin{document}

\title{Texture-aware Intrinsic Image Decomposition with Model- and Learning-based Priors\thanks{* Corresponding author: Xin Yuan.}}

\author{
Xiaodong Wang\textsuperscript{1,2},
Zijun He\textsuperscript{1,2},
Xin Yuan\textsuperscript{2,*} \\
\textsuperscript{1}Zhejiang University, Hangzhou, Zhejiang, China \\
\textsuperscript{2}School of Engineering, Westlake University, Hangzhou, Zhejiang, China
}

\maketitle

\begin{abstract}
This paper aims to recover the intrinsic reflectance layer and shading layer given a single image. Though this intrinsic image decomposition problem has been studied for decades, it remains a significant challenge in cases of complex scenes, i.e. spatially-varying lighting effect and rich textures. In this paper, we propose a novel method for handling severe lighting and rich textures in intrinsic image decomposition, which enables to produce high-quality intrinsic images for real-world images. Specifically, we observe that previous learning-based methods tend to produce texture-less and over-smoothing intrinsic images, which can be used to infer the lighting and texture information given a RGB image. In this way, we design a texture-guided regularization term and formulate the decomposition problem into an optimization framework, to separate the material textures and lighting effect. We demonstrate that combining the novel texture-aware prior can produce superior results to existing approaches. Code is available at \url{https://github.com/xiaodongwo/Efficient_IID}.
\end{abstract}

\begin{IEEEkeywords}
intrinsic image decomposition, structure-texture separation
\end{IEEEkeywords}

\section{Introduction}
\label{sec:intro}

Intrinsic image decomposition is a fundamental mid-level vision problem that aims to separate an image $\Imat$ into its reflectance layer $\Rmat$ and shading layer $\Smat$ as
$\Imat = \Rmat \odot \Smat$, where ``$\odot$" denotes element-wise multiplication. Shading image contains information about illumination conditions and scene geometry, whereas the reflectance image captures the material's reflectance properties, which remain invariant to lighting conditions and shadow effects. Intrinsic image decomposition has been a focal point of research due to its wide applications in computer vision and computer graphics, such as relighting and material manipulation. Moreover, it plays a crucial role in low-light image enhancement and color correction.

Intrinsic image decomposition is inherently an ill-posed problem since there are infinite pairs of reflectance and shading that lead to the same input. Although considerable advancements have been made, current methods may still fall short when it comes to producing high-quality intrinsic images for real-world scenes. This is because real-world images contain a wide range of materials sometimes with high-frequency textures and unknown complex illumination.

\begin{figure}[!t]
	\centering
	\includegraphics[width=\columnwidth]{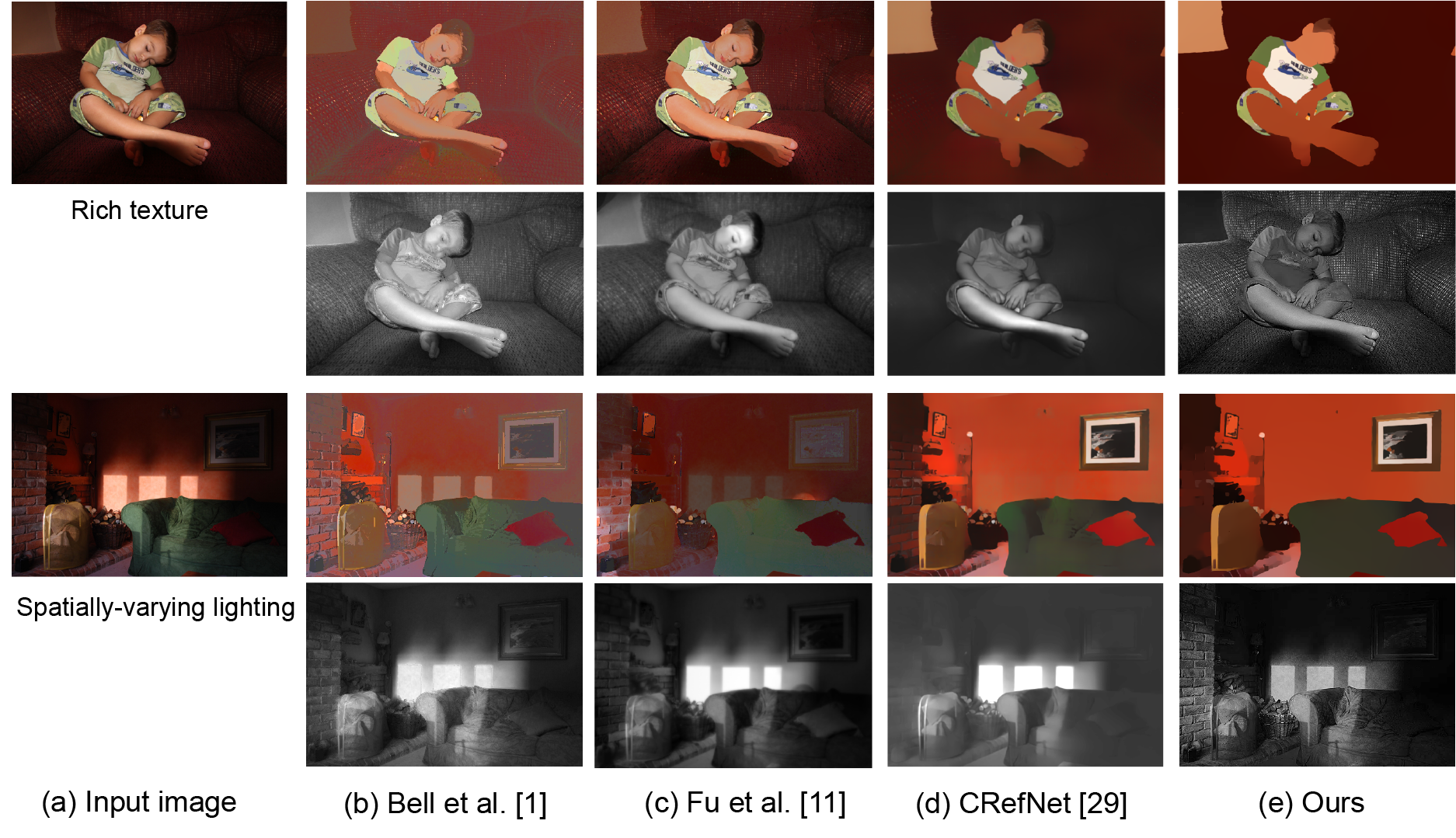}
    \vspace{-5mm}
	\caption{ Comparison between our proposed method with existing state-of-the-art methods on the IIW datasets~\cite{Bell2014In}. Our intrinsic image decomposition method is able to deal with complex scenes with rich textures (top) and spatially-varying lighting conditions (bottom). Please zoom in for details.}
	\label{illustrations}
	\vspace{-2pt}
\end{figure}

Many efforts have been made to address this ill-posed inverse problem. These methods can be roughly divided into two categories: model-based methods~\cite{Bell2014In,Zhao2012A,Gr2009Gr,cheng2019non} and learning-based methods~\cite{narihira2015learning,careaga2024colorful,luo2024intrinsicdiffusion,jin2023estimating}. Model-based methods use hand-crafted priors to regularize the feasible solutions of reflectance and shading, which is flexible but may fail to generalize to natural images in complex scenes such as spatially-varying lighting conditions. Learning-based methods learns the mapping function between an RGB image (input) and their intrinsic image pairs. During the inference stage, the learned network parameters are fixed so the reconstruction speed is fast. Although the performance of these algorithms is superior, the application is hampered by the cost of training such sophisticated networks and computing capability. Moreover, learning-based methods tend to low-frequency structure information rather than high-frequency texture information.

Bearing the above concerns, in this paper, we propose an efficient method to recover high-quality intrinsic images in the wild. Our method is built upon two observations. $i)$ Firstly, model-based methods fail to disentangle the spatially-varying lighting and rich textures, which present both features in recovered reflectance and shading layers, as is shown in Fig.~\ref{illustrations}(b-c). $ii)$ On the other hand, learning-based method is able to separate the lighting effect from reflectance layer but ends up with missing textures in intrinsic images, as shown in Fig.~\ref{illustrations}(d). High-frequency textures are hardly learned due to the low-frequency bias of neural network. To address these challenges, we present a novel optimization framework incorporating a pre-extracted learning-based prior to separate the spatially-varying lighting and high-frequency textures in intrinsic images. Specifically, the lighting-free reflectance layer is firstly pre-extracted from an off-the-shelf pretrained intrinsic images network. We then leverage $\mathbf{\ell}_0$ norm and total variation (TV) prior based on the pre-extracted reflectance layer for ultimate intrinsic image decomposition. This optimization problem is then solved via the alternating direction of multipliers (ADMM) techniques~\cite{boyd2011distributed}. Our approach generalizes the traditional optimization framework to complex scenes recovery in intrinsic image decomposition. 
The main contributions of this work are summarized as:
\begin{itemize}
  \item We design a {\bf learning-based prior} to regularize the reflectance and shading term in traditional optimization framework, which successfully remove the complex illumination in reflectance and preserve rich material textures in shading. 
  \item We derive an optimization procedure based on the ADMM framework and it leads to excellent results.
  \item The proposed method is quite fast. For an image with 512\texttimes384 pixels, our method takes less than 3 seconds on average to get good results.
\end{itemize}

\begin{figure}[]
	\includegraphics[height = 0.47\columnwidth]{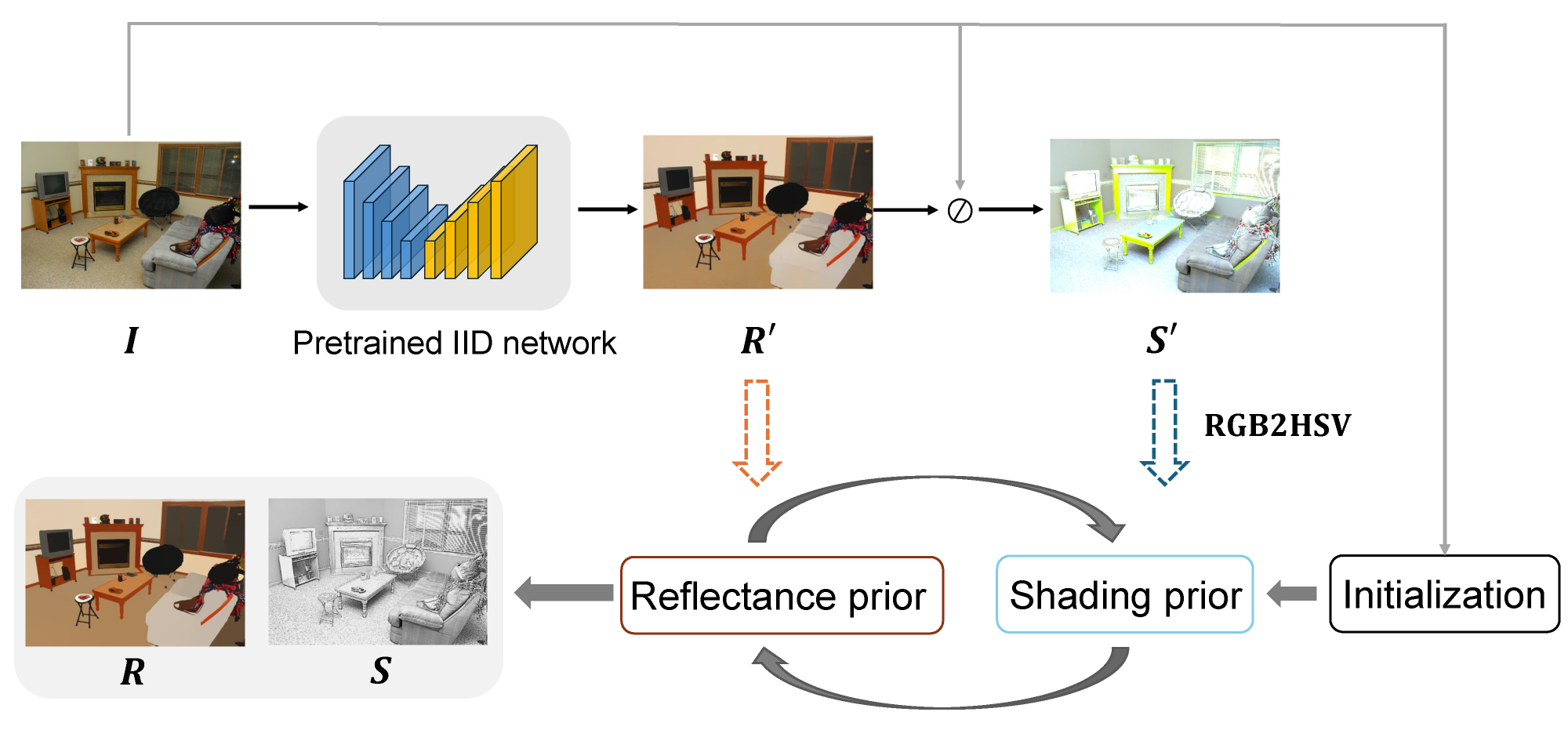}
	\vspace{-10pt}  
	\caption{Flowchart of the proposed optimization algorithm.}
	\label{Pipeline}
\end{figure}

\section{Related Work}

In this section, we review previous works on intrinsic image decomposition, which can be classified into model-based and learning-based methods.

\noindent {\bf Model-based methods:} To address the challenges of intrinsic image decomposition, various priors have been employed. These include grayscale shading~\cite{Bell2014In,Zhao2012A,Gr2009Gr}, smooth shading~\cite{fu2016weighted,fu2019towards}, Retinex theory~\cite{ Zhao2012A, Xu2020STAR, Gr2009Gr}, and sparse reflectance~\cite{fu2019towards,fu2016weighted,rother2011recovering,Bell2014In,garces2012intrinsic,bi20151}. Retinex-based algorithms assume large image gradients are present in reflectance, while smaller gradients stem from shading discontinuities. Reflectance sparsity priors are widely used to enhance spatial coherence in images, either globally or locally. Global sparsity assumes that a scene contains only a limited number of distinct materials, leading to similar reflectance values for groups of pixels. This approach often uses clustering techniques to group pixels with similar properties~\cite{rother2011recovering,Bell2014In,garces2012intrinsic}. Local sparsity, on the other hand, assumes that reflectance is piecewise constant within small regions of the image. This method aims to flatten reflectance maps locally by applying a smoothness constraint~\cite{bi20151,fu2019towards,fu2016weighted}. By doing so, it helps maintain consistent reflectance within homogeneous areas while allowing for variations at object boundaries. Further, many methods have been proposed that require additional input to solve the problem. With user interaction, \cite{Bo2009user} showed that a very high-quality decomposition can be obtained. Researches in~\cite{chen2013simple,jeon2014intrinsic} incorporate additional depth/surface normal input data to provide scene geometry information. The work in~\cite{cheng2019non} introduced an NIR image to provide shading information. For comprehensive reviews of these methods please refer to surveys by Bonneel et al.~\cite{Bon2017Int} and Garces et al.~\cite{garces2022survey}.

\noindent {\bf Learning-based methods:} Data-driven methods have become a popular tool for learning priors and knowledge from training data. Early approaches focused on estimating the relative lightness of reflectance~\cite{narihira2015learning} or ordinal relationships between reflectance values~\cite{zhou2015learning}. Then, various convolutional neural networks have been proposed for dense prediction tasks~\cite{fan2018revisiting,liu2020unsupervised}. Training on real-world datasets such as IIW~\cite{Bell2014In}, providing annotations for similar reflectance properties, has been shown to improve the consistency of reflectance predictions~\cite{li2018cgintrinsics,fan2018revisiting}. Additionally, several techniques have been demonstrated to enhance local reflectance consistency, including generating labels similar to those in the IIW dataset from dense ground truth~\cite{li2018cgintrinsics}, employing an unsupervised reflectance smoothness loss term~\cite{li2018cgintrinsics,li2018learning}, using filters to flatten network predictions~\cite{fan2018revisiting}, disentangling reflectance into flat surfaces and boundaries, and incorporating priors and classifiers to reduce specularities and shadows~\cite{jin2023estimating}. Some methods have also explored diffusion models that jointly learn depth and/or normal prediction with intrinsic image decomposition~\cite{luo2024intrinsicdiffusion,zeng2024rgb}. Learning in an unsupervised fashion from multi-illumination datasets has also been developed~\cite{careaga2023intrinsic}. Besides, several works have used transformer to explore the local characteristics of the intrinsic image decomposition problem~\cite{forsyth2021intrinsic, luo2023crefnet}.

\section{Proposed Method}
In this section, we present our approach for estimating intrinsic images. We begin by reviewing the image formation model, then introduce an efficient solver based on ADMM to address the  optimization problem by our model. Finally, we provide details on the implementation and parameter settings.

\subsection{Image Formation}
Let $\Imat$ represent an input image with pixel values normalized in the range $[0,1]$, and let $\Rmat$ and $\Smat$ denote the reflectance and shading components, respectively. Assuming that the scene captured in the image $\Imat$  follows Lambertian reflectance, and we can express the intrinsic image decomposition problem as follows:
{\small\begin{equation}
\Imat = \Rmat \odot \Smat, \label{eq}
\end{equation}}
where $\Smat$ is a gray image, which is also extendable to multi-channel shading (similar idea with\cite{careaga2024colorful}), and $\Rmat$ is a multi-channel RGB image. We apply the constraint $ \Imat \leq \Smat$. The goal is to estimate $\Smat$ and $\Rmat$ from the input image $\Imat$. 

\begin{figure*}[!t]
	\centering
	\includegraphics[width=0.8\textwidth]{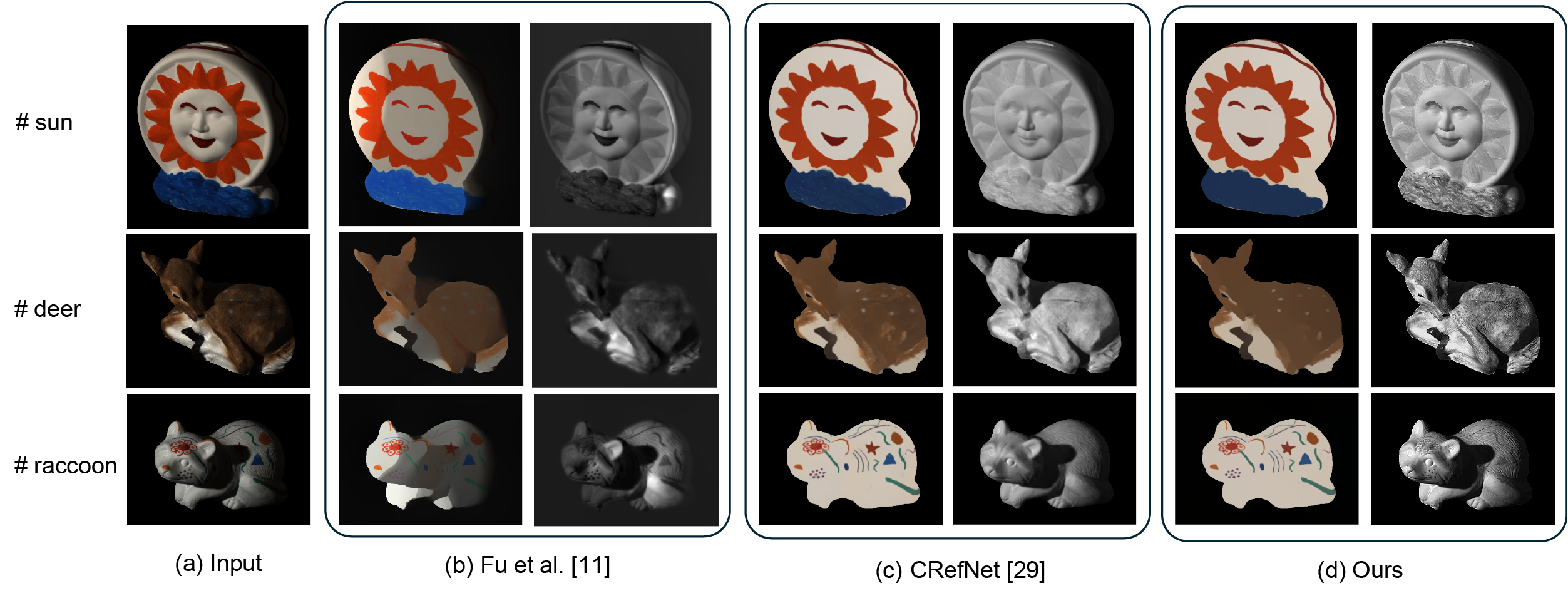}
	\vspace{-10pt}
	\caption{ Comparison between our proposed method with existing state-of-the-art methods on the MIT datasets~\cite{Gr2009Gr}. Please zoom in for details.}
	\label{mit}
	\vspace{-10pt}
\end{figure*}

\subsection{Intrinsic Decomposition Model}
We formulate the intrinsic image decomposition problem as the minimization of the following objective function:
{\small\begin{equation}
E(\Smat, \Rmat) = f_d(\Smat,\Rmat) + \alpha f_r(\Rmat) + \beta f_s(\Smat) + \gamma f_a(\Smat), \label{eq}
\end{equation}}
where $\alpha$, $\beta$, and $\gamma$ are positive parameters for combining three different objective functions. $f_d$ is a data-fidelity term, $f_r$ is the reflectance prior, and $f_s$ is the shading prior; we further add an absolute scale term $f_a$ in the energy function.

\noindent{\bf Data Fidelity:}
Similar to many inverse imaging problems, a data-fidelity term is imposed to regularize the solution. We define the following $\mathbf{\ell}_2$ error metric to quantify the reconstruction error:
{\small
\begin{equation}
f_d = \|\Imat - \Rmat \odot \Smat\|_2^2, \label{eq}
\end{equation}}
where $|\cdot|_p$ denotes the $p$-norm operator.

\noindent{\bf Reflectance Prior:} 
One key observation about real-world images is that reflectance is generally sparse, exhibiting piecewise constant regions. Previous methods adopt $\mathbf{\ell}_1$ \cite{fu2016weighted} or $\mathbf{\ell}_0$ \cite{xu2011image,fu2019towards} norm for sparsity. As reported in \cite{fu2019towards}, $\mathbf{\ell}_0$ norm produce more piecewise constant relectance compared to $\mathbf{\ell}_1$ norm. Moreover, it is less sensitive to nosie and complex shading variations. Therefore we enforce reflectance sparsity by limiting the number of reflectance discontinuities globally, using the $\mathbf{\ell}_0$ norm defined as:
{\small\begin{equation}
f_r = \|\nabla \Rmat\|_0, \label{eq4}
\end{equation}}
where $\nabla$ is the gradient operator. This $\mathbf{\ell}_0$ sparsity term intuitively forces small reflectance discontinuities or noise to zero, while preserving the dominant structures in the reflectance. Further, We introduce a learning-based lighting-free prior into $\mathbf{\ell}_0$ norm such to handle complex lighting scenes, which is introduced in section Model Solver.

\noindent{\bf Shading Prior:} 
Traditional priors involving shading often assume a smooth-varying surface and lighting. While this assumption is effective for simple cases, such as Mondrian-like images with patches of constant reflectance, they often fail in the case of natural images. One key feature of natural images is their complex textures, which can arise from either the reflectance layer (e.g., a flat surface with a dotted pattern) or the shading layer (e.g., a surface with bumps or wrinkles that produce shading patterns). Though researchers have used additional normal map to provide geometry surface hints~\cite{jeon2014intrinsic}, it is still tremendously difficult to tell, even with human annotations. 

We assume that existing intrinsic image decomposition networks (IID network) distinguish and recover the reflectance well. This assumption holds since the neural network is generally more photorealistic compared to traditional optimization methods. On this basis, a direct inversion with $\Imat \oslash\Rmat'$ provides enough material textures of shading, where $\oslash$ denotes element-wise division. We further impose a Total Variation prior (inspired by \cite{chen2023prior}) to enhance the high-frequency shading textures. This total variation-based shading prior term is:
{\small\begin{equation}
f_s = \|\mathbf{TV} (\Smat- \Smat')\|_1. \label{eq5}
\end{equation}}
where $\Smat'$ is the texture prior guided shading extracted by the off-the-self pretrained IID neural network, and we employ CRefNet\cite{luo2023crefnet} in this work. In practice, we use $\Rmat'$ to denote the pre-extracted reflectance layer. 

\noindent{\bf Absolute Scale Term:} 
In addition to the shading smoothness term, we also incorporate an absolute scale constraint on shading, similar to \cite{fu2019towards}, which is formulated as:
{\small\begin{equation}
f_a = \|\Smat - \Smat_0\|_2^2, \label{eq6}
\end{equation}}
where $\Smat_0$  is a reference shading value, which is a constant value for scale constraint. Adopting this scale constraint has two advantages. First, it helps avoid the ambiguity issue from $\Imat = \Rmat \odot \Smat$, since for each decomposition we can enlarge $\Smat$ by any value and reduce $\Rmat$ by some value to get a different valid decomposition. Second, it ensures that there are as few extreme values of shading as possible.

\subsection{Model Solver}
Based on the above discussion, we formulate the ensemble energy function as 
{\small\begin{align}
E(\Smat, \Rmat) &= \|\Rmat \odot \Smat - \Imat\|_2^2 + \alpha \|\mathbf{TV} (\Smat - \Smat')\|_1 \notag \\
& \quad+ \beta \|\nabla \Rmat\|_0 + \gamma \|\Smat - \Smat_0\|_2^2, \quad \text{s.t.} \quad \Imat \leq \Smat,
\label{eq7}
\end{align}}
we assume $\Rmat \in [0,1]$, therefore $\Imat\leq\Smat$ is imposed element-wise. The objective function in Eq. \eqref{eq7} is non-convex because of the $\mathbf{\ell}_0$ and $\mathbf{\ell}_1$ norm regularization. To solve this problem, we apply alternating projection to split the optimization into two subproblems with respect to $\Rmat$ and $\Smat$. We then iterate between reflectance and shading prior until convergence. In particular, for the $k$-th iteration, the subproblems are as follows:
{\small\begin{equation}
\argmin_{\Rmat} \  \|\Rmat \odot \Smat - \Imat\|_2^2 + \beta \|\nabla\Rmat\|_0,
\label{eq8}
\end{equation}
\begin{equation}
\argmin_{\Smat} \  \|\Rmat \odot \Smat - \Imat\|_2^2 + \alpha \|\mathbf{TV} (\Smat - \Smat')\|_1 + \gamma \|\Smat - \Smat_0\|_2^2.
\label{eq9}
\end{equation}}

\noindent{\bf $\Rmat$ subproblem:} we introduce an auxiliary variable $\Gmat$ and solve for $\Rmat$ and $\Gmat$ simultaneously by the minimization below: 
{\small\begin{equation}
\argmin_{\Rmat,\Gmat} \  \|\Rmat - \Imat\oslash\Smat\|_2^2 + \beta \|\Gmat\|_0 + \mu \|\Gmat - \nabla \Rmat\|_2^2 .
\label{eq_RG}
\end{equation}}
As introduced in \cite{xu2011image}, this problem can be further decomposed into two steps estimating $\Rmat$ and $\Gmat$ respectively. $\Rmat$ is computed by diagonalizing operators after Fast Fourier Transform (FFT) for speedup,
{\footnotesize
\begin{equation}
(\Gmat_x, \Gmat_y) = 
\begin{cases} 
(0, 0) & \text{if } (\nabla_x \Rmat)^2 + (\nabla_y \Rmat)^2 \leq \beta/ \mu \\
(\nabla_x \Rmat, \nabla_y \Rmat) & \text{otherwise}
\end{cases},
\label{eq11}
\end{equation}
}
{\footnotesize
\begin{equation}
\Rmat = \mathcal{F}^{-1} \left( \frac{\mathcal{F}(\Imat\oslash\Smat) + \mu (\mathcal{F}(\nabla_x)^* \mathcal{F}(\Gmat_x) + \mathcal{F}(\nabla_y)^* \mathcal{F}(\Gmat_y))}{\mathcal{F}(1) + \mu (\mathcal{F}(\nabla_x)^* \mathcal{F}(\nabla_x) + \mathcal{F}(\nabla_y)^* \mathcal{F}(\nabla_y))} \right)
\label{eq12}
\end{equation}}
where $\mathcal{F}$, $\mathcal{F}^*$ and $\mathcal{F}^{-1}$ denote the FFT operator, its complex conjugate and inverse FFT operator respectively. $\mathcal{F}(1)$ is the Fourier Transform of the delta function. $\nabla_x$ and $\nabla_y$ are the first-order derivative operator along horizontal and vertical direction respectively. In practice, we implement our algorithm by replacing $\mathcal{F}(\Imat\oslash\Smat)$ with a fixed lighting-free reflectance $\mathcal{F}(\Rmat')$, thereby instantly injecting a lighting-free reflectance prior. We use $\mathbf{\ell}_0(\Rmat,\Rmat')$ to denote the imposed $\mathbf{\ell}_0$ norm with pre-extracted prior $\Rmat'$. Fig. \ref{Pipeline} depicts the general pipeline for our proposed optimization algorithm.

\noindent{\bf $\Smat$ subproblem:} we introduce another auxiliary variable $\Hmat$ and solve for $\Smat$ and $\Hmat$ simultaneously by the minimization below: 
{\small\begin{align}
\argmin_{\Smat,\Hmat} \ & \| \Smat - {\Imat}\oslash{\Rmat}\|_2^2 + \alpha \|\mathbf{TV} (\Hmat - \Smat')\|_1 \notag\\
&+ \sigma\|\Hmat-\Smat\|_2^2 + \gamma \|\Smat - \Smat_0\|_2^2, 
\label{eq13}
\end{align}}
where $\sigma$ is a constant parameter. We can alternate between $\Smat$ and $\Hmat$ to gradually approach the solution. Let us consider the following two subproblems:
{\small\begin{align}
\argmin_{\Smat} &\  \| \Smat - {\Imat}\oslash{\Rmat}\|_2^2  + \sigma\|\Hmat-\Smat\|_2^2 + \gamma \|\Smat - \Smat_0\|_2^2,
\label{eq14} \\
\argmin_{\Hmat} &\  \alpha \|\mathbf{TV} (\Hmat - \Smat')\|_1  + \sigma\|\Hmat-\Smat\|_2^2.
\label{eq15}
\end{align}}
Eq. \eqref{eq14} is a quadratic term with close-formed solution. Eq. \eqref{eq15} is a TV denoising term in classical image denoising inverse problems and thus can be denoted as $\textsf{Denoiser}(\Smat, \Smat')$. In practice, we employ a 3D TV denoising algorithm in our experiments, we choose 3D TV (instead of 2D TV) since we want the color variation to be small as possible. Specifically,
{\small\begin{align}
\Smat &= \frac{\Imat\oslash\Rmat+\sigma \Hmat + \gamma \Smat_0}{\sigma+\gamma+1},
\label{eq16}\\
\Hmat &= \textsf{Denoiser}(\Smat, \Smat').
\label{eq17}
\end{align}}
Note that we have incorporated the pre-extracted learning-based prior into Eq. \eqref{eq12} and Eq. \eqref{eq17} for better reconstruction. We then iterate through Eq. \eqref{eq11}, Eq. \eqref{eq12}, Eq. \eqref{eq16}, Eq. \eqref{eq17} to gradually approach the solution.
The whole algorithm is outlined in Algorithm~\ref{alg1}.

\begin{algorithm}
\caption{Intrinsic images using learning-based priors}
\begin{algorithmic}[1]
		 \REQUIRE $\Imat, \Rmat', \Smat', \beta, \mu, \alpha, \Smat_0, \sigma, \gamma$
            \STATE Initial $\Rmat\leftarrow\Imat,\Hmat\leftarrow{\Imat}\oslash{\Rmat}, k=0$.
            \WHILE{Not Converge}
            \STATE  Compute $\Gmat_x, \Gmat_y$ using Eq.~\eqref{eq11};
            \STATE  Update $\Rmat$ using Eq.~\eqref{eq12};
            \STATE  Update $\Smat$ using Eq.~\eqref{eq16};
            \STATE  Update $\Hmat$ using Eq.~\eqref{eq17}.
            \STATE  $k = k+1$.
            \ENDWHILE
            \STATE Return $\Rmat,\Smat$.
\end{algorithmic}
\label{alg1}
\end{algorithm}

\section{Experimental Result}
We implement our proposed algorithm using Matlab on a desktop Intel Core i9 CPU and 128GB RAM. We have experimentally set the parameters as follows: $\mu=1$, $\beta=0.01$, $\sigma=1$, $\gamma=1$, $\alpha=2$, $\Smat_0=0.5$. For initialization, $\Rmat$ is set to be identical to $\Imat$. Meanwhile, $\Smat$ is initialized through summing up the element-wise division of $\Imat$ by $\Rmat$. $\Rmat'$ is pre-extracted from CREfNet \cite{luo2023crefnet}, and $\Smat'$ is obtained derived by converting the image $(\Imat\oslash\Rmat')$ from the RGB color space to the Value component of the HSV color space. $\Smat$ and $\Rmat$ are updated until $\varepsilon_\Xmat = (\|\Xmat^k - \Xmat^{k-1}\| \oslash \|\Xmat^{k-1}\|) \leq \varepsilon$, where $\Xmat$ could be $\Rmat$ or $\Smat$. $\varepsilon$ is set to be 0.00001. Surprisingly, we found that it converges within 3 iterations in our experiments.

\begin{figure}[]
	\includegraphics[width = \columnwidth]{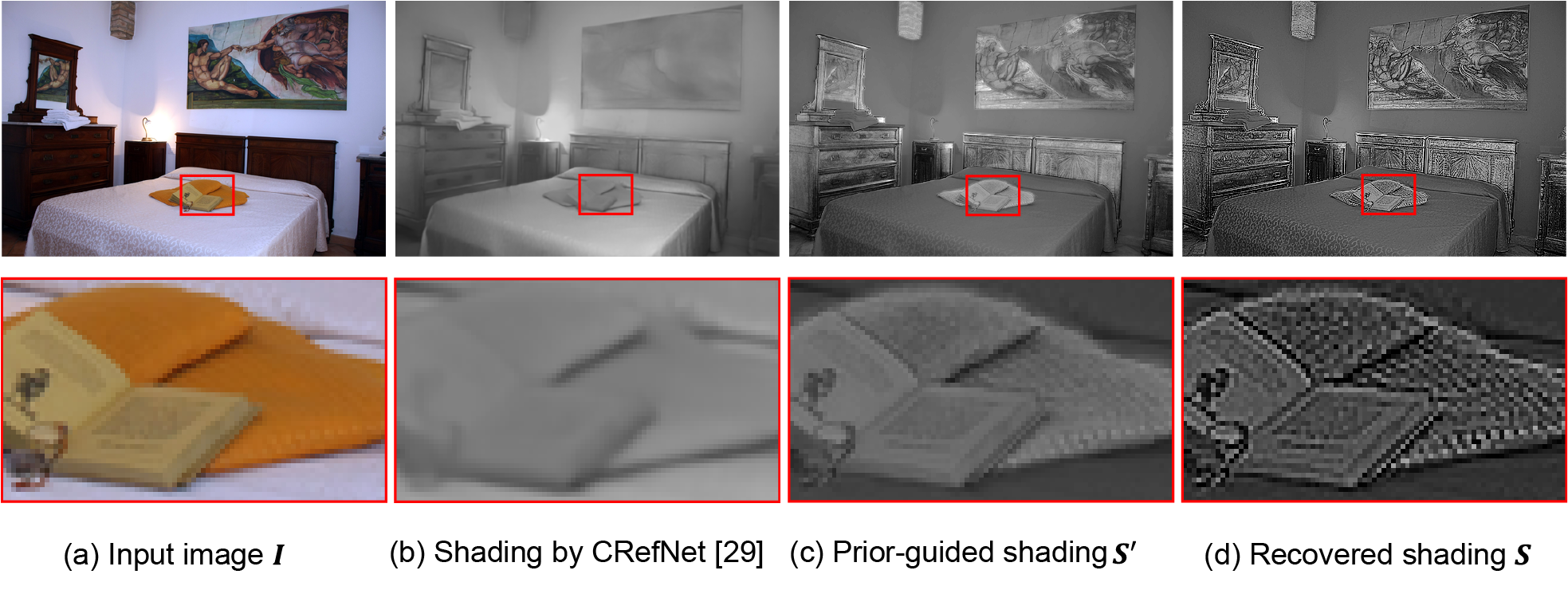}
	\vspace{-10pt}  
	\caption{Enhanced texture in shading using our proposed TV prior.}
	\label{ShadingPrior}
\end{figure}

\subsection{ Comparison with State-of-the-art Methods}\label{CSM}
We evaluate the performance of our proposed method on the IIW dataset~\cite{Bell2014In}, which contains 5230 images spanning a diverse set of scenes, each annotated with approximately 100 human judgments on reflectance as ground truth. We employ a commonly used WHDR metrics, and a lower WHDR value indicates better performance. We compare our method with model-based methods:
Grosse \textit{et al.} \cite{Gr2009Gr}, Bell \textit{et al.} \cite{Bell2014In}, Zhao \textit{et al.} \cite{Zhao2012A}, Fu \textit{et al.} \cite{fu2019towards}, and learning-based methods: Li \textit{et al.} \cite{li2018learning}, NIID-Net \cite{luo2020niid}, CRefNet \cite{luo2023crefnet}. The results are compared with the parameters recommended in the papers. Table \ref{SOTA} demonstrates the quantitative comparison results. We can see that our method outperforms other methods in terms of WHDR. Note that with fine-tuned CRefNet \cite{luo2023crefnet}, the average WHDR drops to 10.8\%, but this requires heavy computational resource and training time, whereas our method offers an efficient and fast way for intrinsic image decomposition, along with enhanced material textures. For visualization comparison, as shown in Fig. \ref{illustrations}, our method recovers lighting-free reflectance and texture-enhanced shading images when compared with other methods. We also performed visual comparison on MIT datasets~\cite{Gr2009Gr}, as shown in Fig. \ref{mit}. It can be seen that our method produces texture-enhanced material shading. Note that all outputs are rescaled for display.

\begin{figure}[]
	\includegraphics[width = \columnwidth]{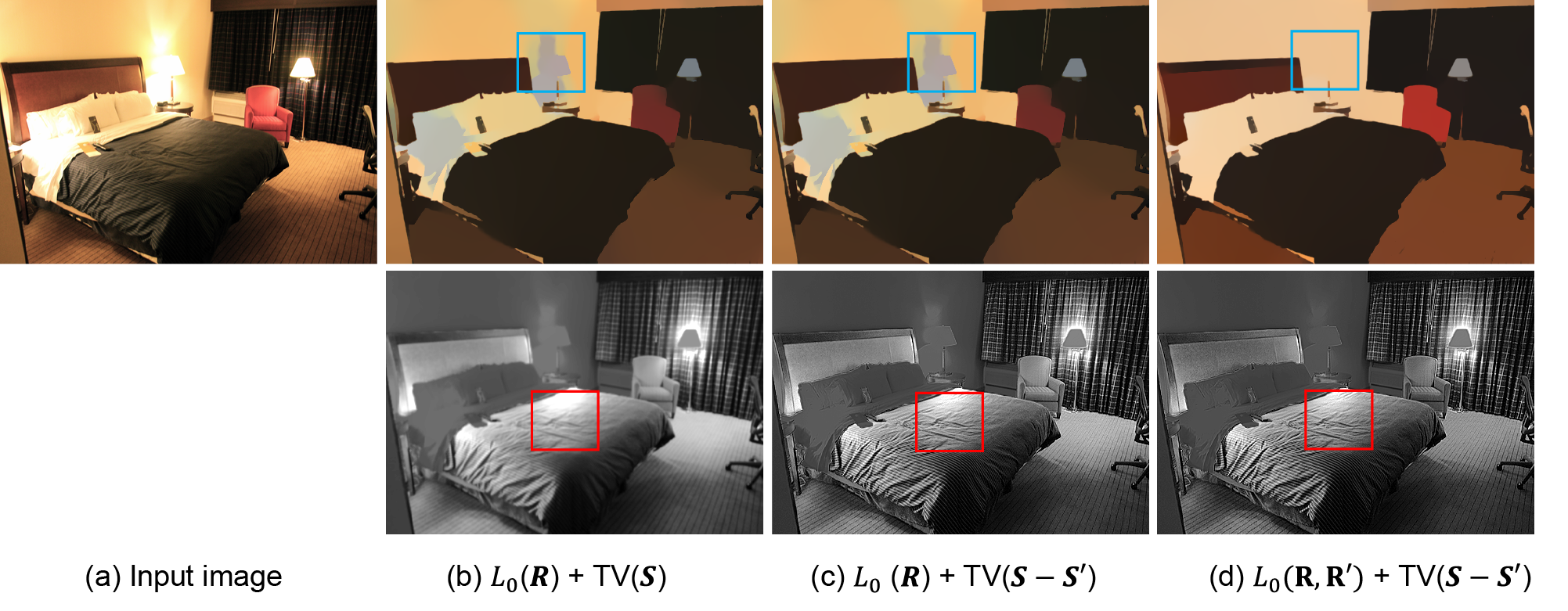}
	\vspace{-10pt}  
	\caption{Demonstration of shading and reflectance prior.}
	\label{DualPrior}
\end{figure}

\begin{table}[htbp]
\centering
\caption{Quantitative comparison between our method and the state-of-the-art methods on the IIW datasets.}
\label{SOTA}
\begin{tabular}{l|cc}
\hlineB{3}
\textbf{Method} & \textbf{WHDR (IIW)} & \textbf{Running time (512\texttimes384)} \\
\hlineB{3}
Retinex-color \cite{Gr2009Gr} & 26.89\% & 250.30s \\
Zhao \textit{et al.} \cite{Zhao2012A} & 23.20\% & 7.11s \\
Bell \textit{et al.} \cite{Bell2014In} & 20.64\% & 209.00s \\
Fu \textit{et al.} \cite{fu2019towards} & 19.20\% & 9.60s \\
Li \textit{et al.} \cite{li2018cgintrinsics} & 15.40\% & - \\
NIID-Net \cite{luo2020niid} & 16.60\% & - \\
CRefNet \cite{luo2023crefnet} & 14.50\% & - \\
\hlineB{2}
Ours & \textbf{13.62\%} & \textbf{2.27s} \\
\hlineB{3}
\end{tabular}
\end{table}

\subsection{Discussion}
\begin{figure}[]
	\includegraphics[height = 0.4\columnwidth]{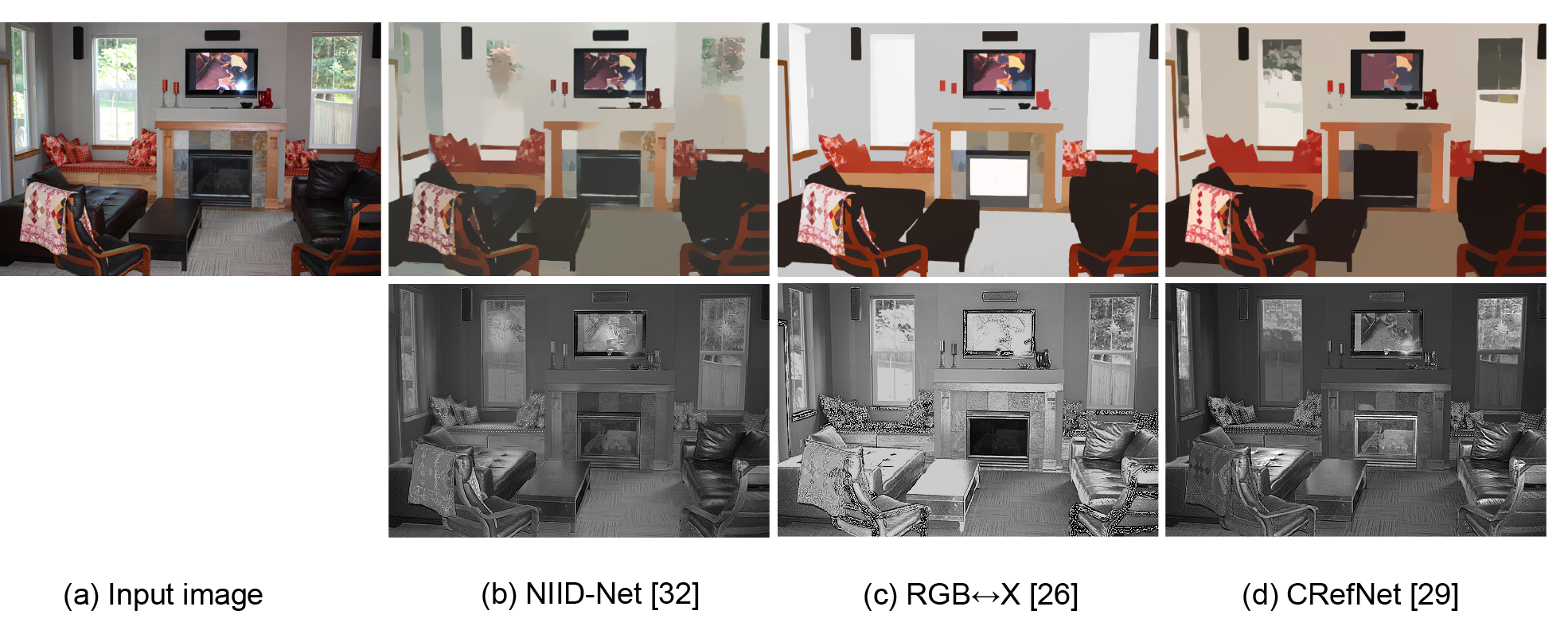}
	\vspace{-10pt}  
	\caption{Demonstration of intrinsic image decomposition using different IID networks as pre-extraction prior.}
	\label{IIDPrior}
\end{figure}

\noindent{\bf Ablation study:}  We demonstrate the effectiveness of our proposed reflectance and shading priors, shown in Fig. \ref{ShadingPrior} and \ref{DualPrior}. As compared with the gray shading proposed in CRefNet and the pre-extracted shading $\Smat'$, our method produces texture-enhanced shading layer. Fig. \ref{DualPrior} demonstrates the lighting-free reflectance and texture-enhanced shading using $\mathbf{\ell}_0$ and 3D TV respectively, where $\mathbf{\ell}_0(\Rmat,\Rmat')$ represents the $\mathbf{\ell}_0$ norm incorporating the learning-based reflectance prior $\Rmat'$, $\mathbf{TV}(\Smat-\Smat')$ represents the 3D TV prior with pre-extracted shading prior $\Smat'$. Moreover, we have compared our methods using different pre-extracted results from different intrinsic image decomposition networks: NIID-Net \cite{luo2020niid}, RGB2X \cite{zeng2024rgb} and CRefNet \cite{luo2023crefnet}, as shown in Fig. \ref{IIDPrior}. We choose CRefNet as our baseline for pre-extracted prior because it exhibits a greater visual reflectance shown in the top row in Fig. \ref{IIDPrior}. Though the recovered reflectance and shading layer vary from networks, the general features are consistent with our conclusion that the shading shows enhanced textures whereas the reflectance is free from spatially-varying lighting and shadow effects.

\noindent{\bf Running time:} 
We compare the average running times of different methods on the images in Table \ref{SOTA}. All the experiments are conducted on a desktop with Intel Core i9 CPU and 128GB RAM. The image size is 512\texttimes384. Note that no learning-based methods are included in this table since we only test the data in inference stage, which is negligible compared with running time on optimization methods. In summary, our method is fast without loss of accuracy.

\begin{figure}[]
    \centering
	\includegraphics[height = 0.45\columnwidth]{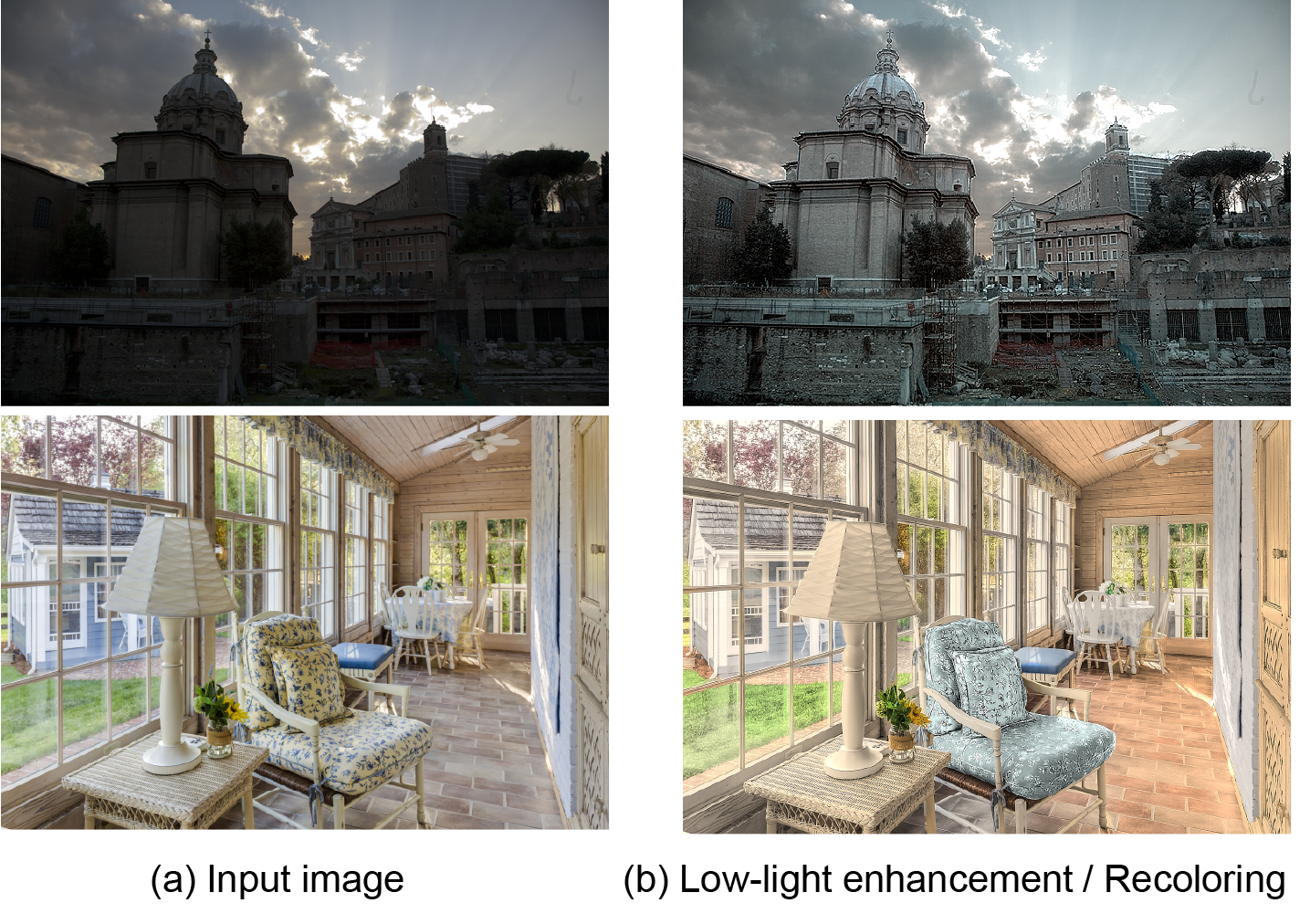}
	\vspace{-10pt}  
	\caption{Application to low-light image enhancement and image recoloring.}
	\label{Applica}
\end{figure}

\noindent{\bf Applications to low-light enhancement and recoloring:} 
We apply our method to low-light image enhancement and image recoloring. As shown in Fig. \ref{Applica}, given a low-light image, we firstly estimate the reflectance and shading layer using our proposed method, and apply a gamma correction ($\gamma = 2.2$) to the shading layer. We can then obtain the final enhanced image by multiplying this corrected shading layer with estimated reflectance. For image recoloring, we replace the material color (or texture, the chair in Fig. \ref{Applica}) with another material in reflectance layer and compose recolored image (see pipelines in \cite{careaga2024colorful}). Notice that color and high-frequency textures are enhanced in recolored image using our method. We can see that our method generalizes to image enhancement and image editing tasks well.

\noindent{\bf Limitations and future work:} 
The limitation of our work is that it is challenging to deal with textures with colors. Our method is robust to shadows and spatial-varying lighting but may fail to distinguish color textures in reflectance layer. This is because the $\mathbf{\ell}_0$ norm tends to oversmooth the reflectance layer and thus mis-distribute the color texture to shading texture layer. Overall our method could be a potential tool to incorporate the learning-based prior to traditional optimization framework and offers a robust way for intrinsic image decomposition, structure-texture decomposition, image editing, low-light image enhancement and so on.

\section*{Conclusion}

We have presented an efficient method to recover
high-quality intrinsic images for real-world scenes. Our method incorporate a lighting-free prior (pre-extracted from an intrinsic image decomposition network) into a traditional optimization framework, thus enabling intrinsic images with lighting-free and texture enhancement decomposition.  We have introduced a $\mathbf{\ell}_0$ norm to impose the sparsity and lighting-free features of the reflectance, and designed a novel denoising term to enhance the material texture of the shading layer. We have performed experiments on IIW dataset and MIT dataset, thus demonstrating its feasibility. We want to emphasize that our methods are generally suitable for complex scenes with spatially-varying lighting and high-frequency material textures. For spectrally-varying textures it will be challenging to distinguish, and this will be our future work.

\section*{Acknowledgment}

This work was supported by National Key R\&D Program of China (2024YFF0505603), the National Natural Science Foundation of China (grant number 62271414), Zhejiang Provincial Distinguished Young Scientist Foundation (grant number LR23F010001), Zhejiang “Pioneer” and “Leading Goose” R\&D Program (grant number 2024SDXHDX0006, 2024C03182), the Key Project of Westlake Institute for Optoelectronics (grant number 2023GD007), the 2023 International Sci-tech Cooperation Projects under the purview of the “Innovation Yongjiang 2035” Key R\&D Program (grant number 2024Z126).

{\footnotesize
\bibliographystyle{IEEEbib}
\bibliography{icme2025references}

\begin{thebibliography}{10}

\bibitem{Bell2014In}
Sean Bell, Kavita Bala, and Noah Snavely,
\newblock ``Intrinsic images in the wild,''
\newblock {\em ACM Trans. Graph.}, vol. 33, no. 4, July 2014.

\bibitem{Zhao2012A}
Qi~Zhao, Ping Tan, Qiang Dai, Li~Shen, Enhua Wu, and Stephen Lin,
\newblock ``A closed-form solution to retinex with nonlocal texture constraints,''
\newblock {\em IEEE Transactions on Pattern Analysis and Machine Intelligence}, vol. 34, no. 7, pp. 1437--1444, 2012.

\bibitem{Gr2009Gr}
Roger Grosse, Micah~K. Johnson, Edward~H. Adelson, and William~T. Freeman,
\newblock ``Ground truth dataset and baseline evaluations for intrinsic image algorithms,''
\newblock in {\em 2009 IEEE 12th International Conference on Computer Vision}, 2009, pp. 2335--2342.

\bibitem{cheng2019non}
Ziang Cheng, Yinqiang Zheng, Shaodi You, and Imari Sato,
\newblock ``Non-local intrinsic decomposition with near-infrared priors,''
\newblock in {\em Proceedings of the IEEE/CVF international conference on computer vision}, 2019, pp. 2521--2530.

\bibitem{narihira2015learning}
Takuya Narihira, Michael Maire, and Stella~X Yu,
\newblock ``Learning lightness from human judgement on relative reflectance,''
\newblock in {\em Proceedings of the IEEE conference on computer vision and pattern recognition}, 2015, pp. 2965--2973.

\bibitem{careaga2024colorful}
Chris Careaga and Ya{\u{g}}{\i}z Aksoy,
\newblock ``Colorful diffuse intrinsic image decomposition in the wild,''
\newblock {\em ACM Transactions on Graphics (TOG)}, vol. 43, no. 6, pp. 1--12, 2024.

\bibitem{luo2024intrinsicdiffusion}
Jundan Luo, Duygu Ceylan, Jae~Shin Yoon, Nanxuan Zhao, Julien Philip, Anna Fr{\"u}hst{\"u}ck, Wenbin Li, Christian Richardt, and Tuanfeng Wang,
\newblock ``Intrinsicdiffusion: joint intrinsic layers from latent diffusion models,''
\newblock in {\em ACM SIGGRAPH 2024 Conference Papers}, 2024, pp. 1--11.

\bibitem{jin2023estimating}
Yeying Jin, Ruoteng Li, Wenhan Yang, and Robby~T Tan,
\newblock ``Estimating reflectance layer from a single image: Integrating reflectance guidance and shadow/specular aware learning,''
\newblock in {\em Proceedings of the AAAI Conference on Artificial Intelligence}, 2023, vol.~37, pp. 1069--1077.

\bibitem{boyd2011distributed}
Stephen Boyd, Neal Parikh, Eric Chu, Borja Peleato, Jonathan Eckstein, et~al.,
\newblock ``Distributed optimization and statistical learning via the alternating direction method of multipliers,''
\newblock {\em Foundations and Trends{\textregistered} in Machine learning}, vol. 3, no. 1, pp. 1--122, 2011.

\bibitem{fu2016weighted}
Xueyang Fu, Delu Zeng, Yue Huang, Xiao-Ping Zhang, and Xinghao Ding,
\newblock ``A weighted variational model for simultaneous reflectance and illumination estimation,''
\newblock in {\em Proceedings of the IEEE conference on computer vision and pattern recognition}, 2016, pp. 2782--2790.

\bibitem{fu2019towards}
Gang Fu, Qing Zhang, and Chunxia Xiao,
\newblock ``Towards high-quality intrinsic images in the wild,''
\newblock in {\em 2019 IEEE International Conference on Multimedia and Expo (ICME)}. IEEE, 2019, pp. 175--180.

\bibitem{Xu2020STAR}
Jun Xu, Yingkun Hou, Dongwei Ren, Li~Liu, Fan Zhu, Mengyang Yu, Haoqian Wang, and Ling Shao,
\newblock ``Star: A structure and texture aware retinex model,''
\newblock {\em IEEE Transactions on Image Processing}, vol. 29, pp. 5022--5037, 2020.

\bibitem{rother2011recovering}
Carsten Rother, Martin Kiefel, Lumin Zhang, Bernhard Sch{\"o}lkopf, and Peter Gehler,
\newblock ``Recovering intrinsic images with a global sparsity prior on reflectance,''
\newblock {\em Advances in neural information processing systems}, vol. 24, 2011.

\bibitem{garces2012intrinsic}
Elena Garces, Adolfo Munoz, Jorge Lopez-Moreno, and Diego Gutierrez,
\newblock ``Intrinsic images by clustering,''
\newblock in {\em Computer graphics forum}. Wiley Online Library, 2012, vol.~31, pp. 1415--1424.

\bibitem{bi20151}
Sai Bi, Xiaoguang Han, and Yizhou Yu,
\newblock ``An l 1 image transform for edge-preserving smoothing and scene-level intrinsic decomposition,''
\newblock {\em ACM Transactions on Graphics (TOG)}, vol. 34, no. 4, pp. 1--12, 2015.

\bibitem{Bo2009user}
Adrien Bousseau, Sylvain Paris, and Fr\'{e}do Durand,
\newblock ``User-assisted intrinsic images,''
\newblock {\em ACM Trans. Graph.}, vol. 28, no. 5, pp. 1–10, Dec. 2009.

\bibitem{chen2013simple}
Qifeng Chen and Vladlen Koltun,
\newblock ``A simple model for intrinsic image decomposition with depth cues,''
\newblock in {\em Proceedings of the IEEE international conference on computer vision}, 2013, pp. 241--248.

\bibitem{jeon2014intrinsic}
Junho Jeon, Sunghyun Cho, Xin Tong, and Seungyong Lee,
\newblock ``Intrinsic image decomposition using structure-texture separation and surface normals,''
\newblock in {\em Computer Vision--ECCV 2014: 13th European Conference, Zurich, Switzerland, September 6-12, 2014, Proceedings, Part VII 13}. Springer, 2014, pp. 218--233.

\bibitem{Bon2017Int}
Nicolas Bonneel, Balazs Kovacs, Sylvain Paris, and Kavita Bala,
\newblock ``Intrinsic decompositions for image editing,''
\newblock {\em Computer Graphics Forum}, vol. 36, no. 2, pp. 593--609, 2017.

\bibitem{garces2022survey}
Elena Garces, Carlos Rodriguez-Pardo, Dan Casas, and Jorge Lopez-Moreno,
\newblock ``A survey on intrinsic images: Delving deep into lambert and beyond,''
\newblock {\em International Journal of Computer Vision}, vol. 130, no. 3, pp. 836--868, 2022.

\bibitem{zhou2015learning}
Tinghui Zhou, Philipp Krahenbuhl, and Alexei~A Efros,
\newblock ``Learning data-driven reflectance priors for intrinsic image decomposition,''
\newblock in {\em Proceedings of the IEEE international conference on computer vision}, 2015, pp. 3469--3477.

\bibitem{fan2018revisiting}
Qingnan Fan, Jiaolong Yang, Gang Hua, Baoquan Chen, and David Wipf,
\newblock ``Revisiting deep intrinsic image decompositions,''
\newblock in {\em Proceedings of the IEEE conference on computer vision and pattern recognition}, 2018, pp. 8944--8952.

\bibitem{liu2020unsupervised}
Yunfei Liu, Yu~Li, Shaodi You, and Feng Lu,
\newblock ``Unsupervised learning for intrinsic image decomposition from a single image,''
\newblock in {\em Proceedings of the IEEE/CVF conference on computer vision and pattern recognition}, 2020, pp. 3248--3257.

\bibitem{li2018cgintrinsics}
Zhengqi Li and Noah Snavely,
\newblock ``Cgintrinsics: Better intrinsic image decomposition through physically-based rendering,''
\newblock in {\em Proceedings of the European conference on computer vision (ECCV)}, 2018, pp. 371--387.

\bibitem{li2018learning}
Zhengqi Li and Noah Snavely,
\newblock ``Learning intrinsic image decomposition from watching the world,''
\newblock in {\em Proceedings of the IEEE conference on computer vision and pattern recognition}, 2018, pp. 9039--9048.

\bibitem{zeng2024rgb}
Zheng Zeng, Valentin Deschaintre, Iliyan Georgiev, Yannick Hold-Geoffroy, Yiwei Hu, Fujun Luan, Ling-Qi Yan, and Milo{\v{s}} Ha{\v{s}}an,
\newblock ``Rgb$\leftrightarrow$ x: Image decomposition and synthesis using material-and lighting-aware diffusion models,''
\newblock in {\em ACM SIGGRAPH 2024 Conference Papers}, 2024, pp. 1--11.

\bibitem{careaga2023intrinsic}
Chris Careaga and Ya{\u{g}}{\i}z Aksoy,
\newblock ``Intrinsic image decomposition via ordinal shading,''
\newblock {\em ACM Transactions on Graphics}, vol. 43, no. 1, pp. 1--24, 2023.

\bibitem{forsyth2021intrinsic}
David Forsyth and Jason~J Rock,
\newblock ``Intrinsic image decomposition using paradigms,''
\newblock {\em IEEE transactions on pattern analysis and machine intelligence}, vol. 44, no. 11, pp. 7624--7637, 2021.

\bibitem{luo2023crefnet}
Jundan Luo, Nanxuan Zhao, Wenbin Li, and Christian Richardt,
\newblock ``Crefnet: Learning consistent reflectance estimation with a decoder-sharing transformer,''
\newblock {\em IEEE Transactions on Visualization and Computer Graphics}, 2023.

\bibitem{xu2011image}
Li~Xu, Cewu Lu, Yi~Xu, and Jiaya Jia,
\newblock ``Image smoothing via l 0 gradient minimization,''
\newblock in {\em Proceedings of the 2011 SIGGRAPH Asia conference}, 2011, pp. 1--12.

\bibitem{chen2023prior}
Yurong Chen, Yaonan Wang, and Hui Zhang,
\newblock ``Prior image guided snapshot compressive spectral imaging,''
\newblock {\em IEEE Transactions on Pattern Analysis and Machine Intelligence}, vol. 45, no. 9, pp. 11096--11107, 2023.

\bibitem{luo2020niid}
Jundan Luo, Zhaoyang Huang, Yijin Li, Xiaowei Zhou, Guofeng Zhang, and Hujun Bao,
\newblock ``Niid-net: Adapting surface normal knowledge for intrinsic image decomposition in indoor scenes,''
\newblock {\em IEEE Transactions on Visualization and Computer Graphics}, vol. 26, no. 12, pp. 3434--3445, 2020.

\end{thebibliography}
}

\end{document}